\pdfoutput=1

\documentclass{article}
\usepackage{arxiv}

\usepackage[utf8]{inputenc} 
\usepackage[round]{natbib}
\usepackage[T1]{fontenc}    
\usepackage{hyperref}       
\usepackage{url}            
\usepackage{float}            
\usepackage{booktabs}       
\usepackage{amsfonts}       
\usepackage{nicefrac}       
\usepackage{microtype}      
\usepackage{graphicx}
\usepackage{amsmath} 
\usepackage{cleveref} 
\usepackage{placeins} 
\usepackage[page]{appendix}

\newcommand{\bsr}[0]{(A)}
\newcommand{\ssr}[0]{(B)}

\usepackage{hyperref}

\title{The Two Regimes of Deep Network Training}

\author{
    Guillaume Leclerc \\
	MIT\\
	\texttt{leclerc@mit.edu} \\
	\And
    Aleksander Madry \\
	MIT\\
	\texttt{madry@mit.edu} \\
}

\begin{document}
\twocolumn[
\maketitle
\begin{abstract}
    Learning rate schedule has a major impact on the performance of deep learning
models. Still, the choice of a schedule is often heuristical. We aim to develop
a precise understanding of the effects of different learning rate schedules
and the appropriate way to select them. To this end, we isolate two distinct
phases of training---the first, which we refer to as the ``large-step'' regime, exhibits
a rather poor performance from an optimization point of view but is the primary contributor to
model generalization; the latter, ``small-step'' regime
exhibits much more ``convex-like'' optimization behavior but used in isolation produces models that generalize poorly. We find that by treating these regimes separately---and {\em specializing} our training algorithm to each one of them---we can significantly simplify learning rate schedules.

\vspace{1cm}
\end{abstract}
]

\section{Introduction} \label{sec:introduction}

Finding the right learning rate schedule is critical to obtain the best testing
accuracy for a given neural network architecture. As deep learning started
gaining popularity, starting with the largest learning rate and gradually
decreasing it became the standard practice~\citep{bengio2012practical}. Indeed,
today, such ``step schedule'' still remains one the most popular learning rate
schedules, and when properly tuned it yields competitive models.

More recently, as architectures grew deeper and wider, and as training massive
datasets became the norm, more elaborate schedules emerged
~\citep{loshchilov17sgdr,Smith17,SmithT17}.  While these schedule have shown
great practical success, it is still unclear why that is the case.
\citet{li2019exponential} even showed that, counter-intuitively, an exponentially
\textit{increasing} schedule could also be effective. In the light of this, it
is time to revisit learning rate schedules and shed some light on why some
perform well and some other don't.

\paragraph{Our contributions}

In this paper, we identify two training regimes: (1) the large-step regime and
(2) the small-step one which correspond usually respectively to the start and
the end of ``step-schedule''. In particular, we examine these regimes through
the lens of optimization and generalization. We find that:

\begin{itemize}
\item In the large-step rate regime, the loss does not decrease consistently at
    each epoch and the final loss value obtained after convergence is much higher
    than when training in the small-step regime. In that latter regime, the reduction of
    the loss is faster and smooth and to large degree matches the intuition
    drawn from the convex optimization literature.

\item In the large-step regime, momentum does {\em not} seem to have a discernible
    benefit. More precisely, we show that we can recover similar loss decrease curves for
    a wide range of different momentum values as long as we make a corresponding
    change in the learning rate. In the small-step regime, however,
    momentum becomes crucial to reaching a good solution quickly.

\item Finally, we leverage this understanding to propose a simple two-stage
    learning rate schedule that achieves state of the art performance on the \texttt{CIFAR-10} and \texttt{ImageNet} datasets. Importantly, in this schedule, each stage uses a \emph{different} algorithm and hyper-parameters. 
\end{itemize}

Our findings suggest that it might be beneficial to depart from viewing deep network training as a single optimization problem and instead to explore using different algorithms for different stages of that process. In particular, second order methods (such as K-FAC~\citep{MartensG2015} and
L-BFGS~\citep{LiuN1989})---that are currently viewed as successful at reducing the number of training iterations but leading to suboptimal generalization performance---might be good candidates for using (solely) in the small-step regime.

\section{Background}
\label{sec:background}

Given a (differentiable) function $f(\theta)$, one of the most popular
techniques for minimizing it is to use the gradient descent method (GD). This
method, starting from an initial solution $w^0$, iteratively updates the
solution $w$ as:

\begin{equation}
w^{t+1} = w^t - \eta \nabla f(w^t),
\end{equation}

where $\eta > 0$ is the \textit{learning rate}.  GD is the most natural and
simple continuous optimization scheme. However, there are a host of its most
advanced variants. One of the prominent ones is momentum gradient descent,
often referred to as the \textit{classic momentum}, or \textit{heavy ball
method}~\citep{polyak1964some}.  It corresponds to an update rule:

\begin{equation}
\begin{split}
\label{eqn:update_CM}
g^0 & = {0}, \\
g^{t+1} & = \mu g^t + \nabla f(w^t),\\
w^{t+1} & = w^t - \eta g^{t+1},
\end{split}
\end{equation}

where $\mu$ is a scalar that controls the momentum accumulation.
There are also other variants of momentum dynamics.
Most prominently,  \textit{Nesterov's accelerated gradient}~\citep{nesterov1983method}
offers a theoretically optimal convergence rate. However, it tends to have poor
behavior in practice due its brittleness.
For that reason, and also because of its immense popularity,
we will focus on the above-mentioned classic momentum dynamics instead.

\section{The Two Learning Regimes}

In this work, we will be interested in isolating two learning regimes:
\begin{itemize}
\item[\bsr] \textbf{``large-step'' regime: }corresponds to the highest learning rate that can be
used without causing divergence, as per~\citet{bengio2012practical}.\\
\item[\ssr] \textbf{``small-step'' regime:} corresponds to the largest learning rate at which
loss is consistently decreasing. (In ~\citet{SmithT17}, the authors
propose an experimental procedure to estimate appropriate learning rates).
\end{itemize}

In carefully tuned step-wise learning rate schedules, the first and last learning
rates usually correspond to the large-step and small-step regimes\footnote{We could not
identify a sharp boundary between these two regimes. Learning rates in between
the two extremes seems to essentially be a mixture of the two behaviors.}.
Our goal is to characterize and understand how these regimes differ---first from an optimization and then from a generalization perspective.

\subsection{Optimization perspective}
\label{sec:optimization}

By examining the evolution of the loss from initialization on
\Cref{fig:losses}, we can note three major differences between the two
regimes:

\begin{figure}[h!]
\centering
\includegraphics[width=\columnwidth]{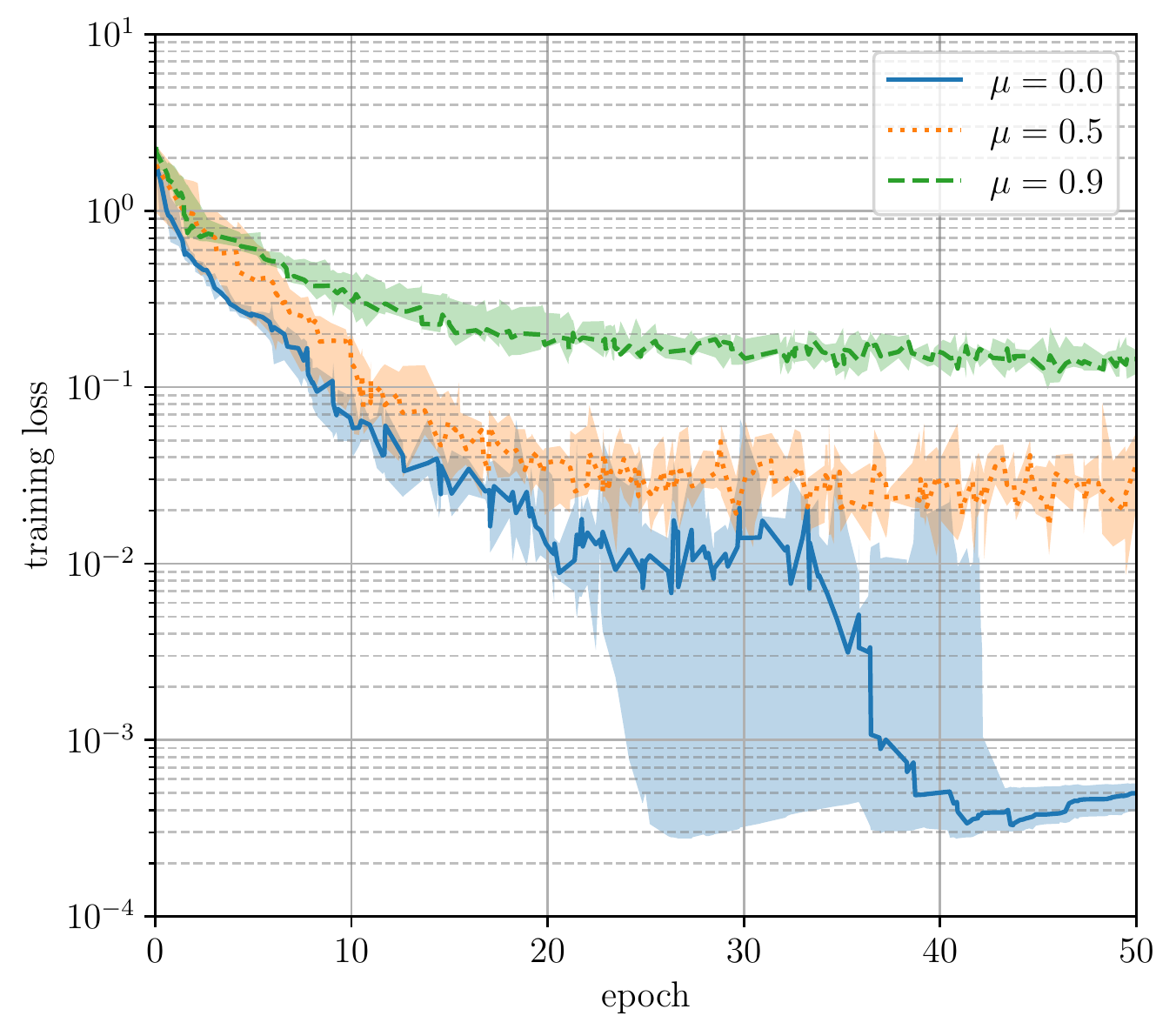}
\includegraphics[width=\columnwidth]{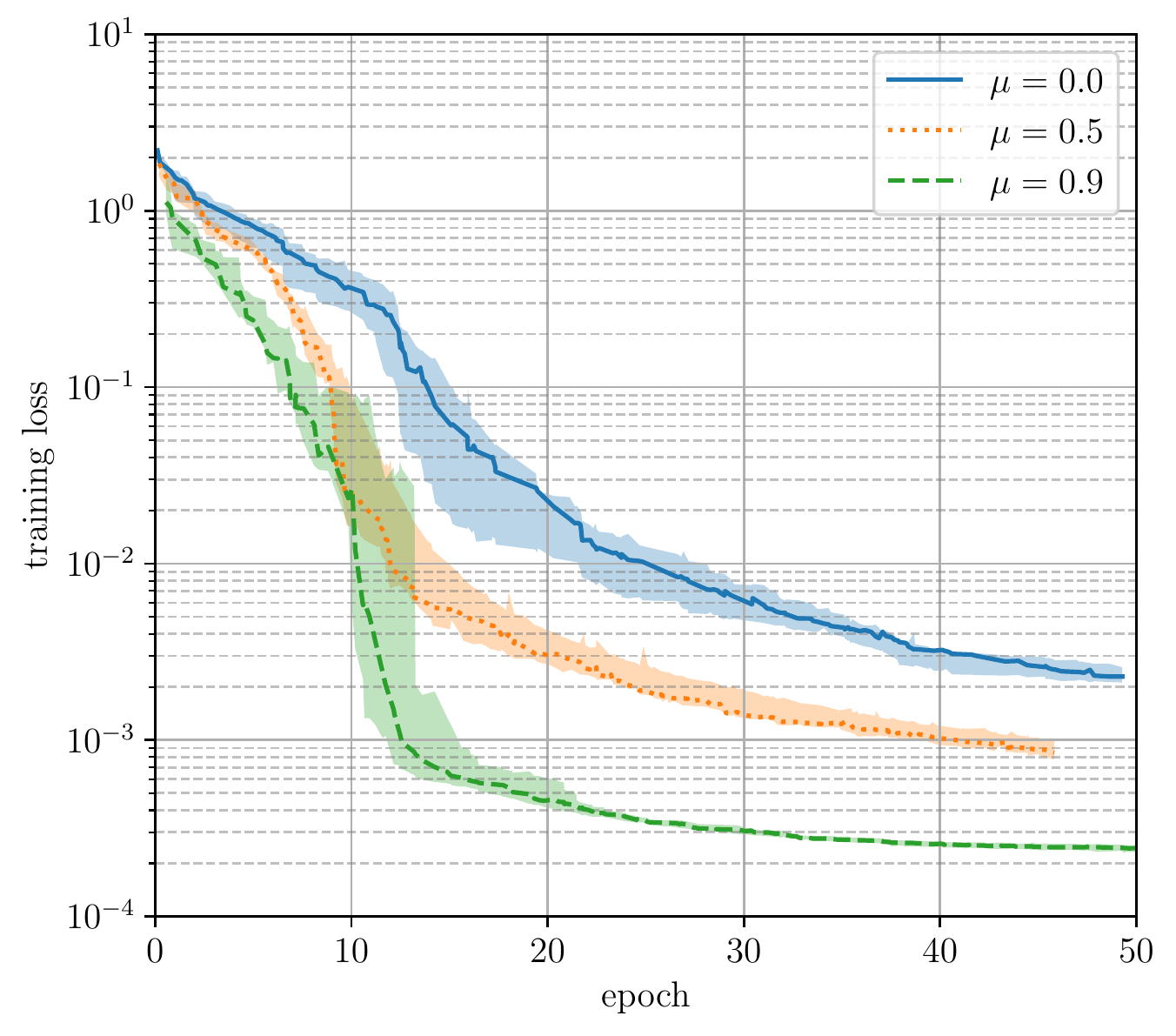}
 \caption{\label{fig:losses}Evolution of the training loss for 50 epochs with
 different momentum values on \texttt{CIFAR-10} and \texttt{VGG-13-BN}. (Top) regime \bsr{} with $\eta=0.1$ and (Bottom) regime \ssr{}  with $\eta=0.001$.}
\end{figure}

\begin{enumerate}
\item The best solution is found in the low learning rate regime, even though we
performed the same number of 100 times smaller steps---which corresponds to a
much shorted distance traveled from the initialization.
\item In regime \bsr{}, the evolution of the loss is very noisy,
    while in \ssr, it decreases almost at each epoch.
\item Momentum seems to behave completely differently in the two experiments. At
    the top of \Cref{fig:losses}, the largest $\mu$ value yields the worst
    solution, whereas on the other, the final loss decreases as we increase
    $\mu$.
\end{enumerate}

These pieces of evidence suggest that regime \bsr{} is a
highly non-convex optimization problem, while the low learning rate regime reflects the intuitions from the convex optimization world. To highlight this
distinction we will use momentum~(as defined in \Cref{sec:background}).

\paragraph{Momentum.} Momentum can provably accelerate gradient descent over functions that are
convex, but does not provide any theoretical guarantees when that property does
not hold. In order to highlight the different nature of the problems we are
solving in each regime, we compare the behavior of momentum when used on a
convex function, and on a deep neural network under both regimes.

Ideally, we would like the momentum vector to be a signal that:
(1) points reliably towards the optimum of our problem, and (2) is strong
enough to actually have an impact on the trajectory. To focus on these two
key properties, we track the two respective quantities:

\begin{enumerate}
    \item \textbf{Alignment}: the angle between the momentum and the direction
        to optimum\footnote{When the optimum is not known, as it is the case for
            neural networks, we use instead the solution our algorithm converged
            to eventually.},
    \begin{equation}s^t = \frac{g^t \cdot (x^* - x^t)}{||g^t||_2||x^* - x^t||_2}.
    \end{equation}
    \item \textbf{Scale}: the ratio between the magnitude of the momentum vector
        and the gradient,
    \begin{equation}r^t = \frac{||g^t||_2}{||\nabla f(w^t)||_2}.
    \end{equation}
    
\end{enumerate}

\begin{figure}[t!]
   \centering
     \includegraphics[width=1\columnwidth]{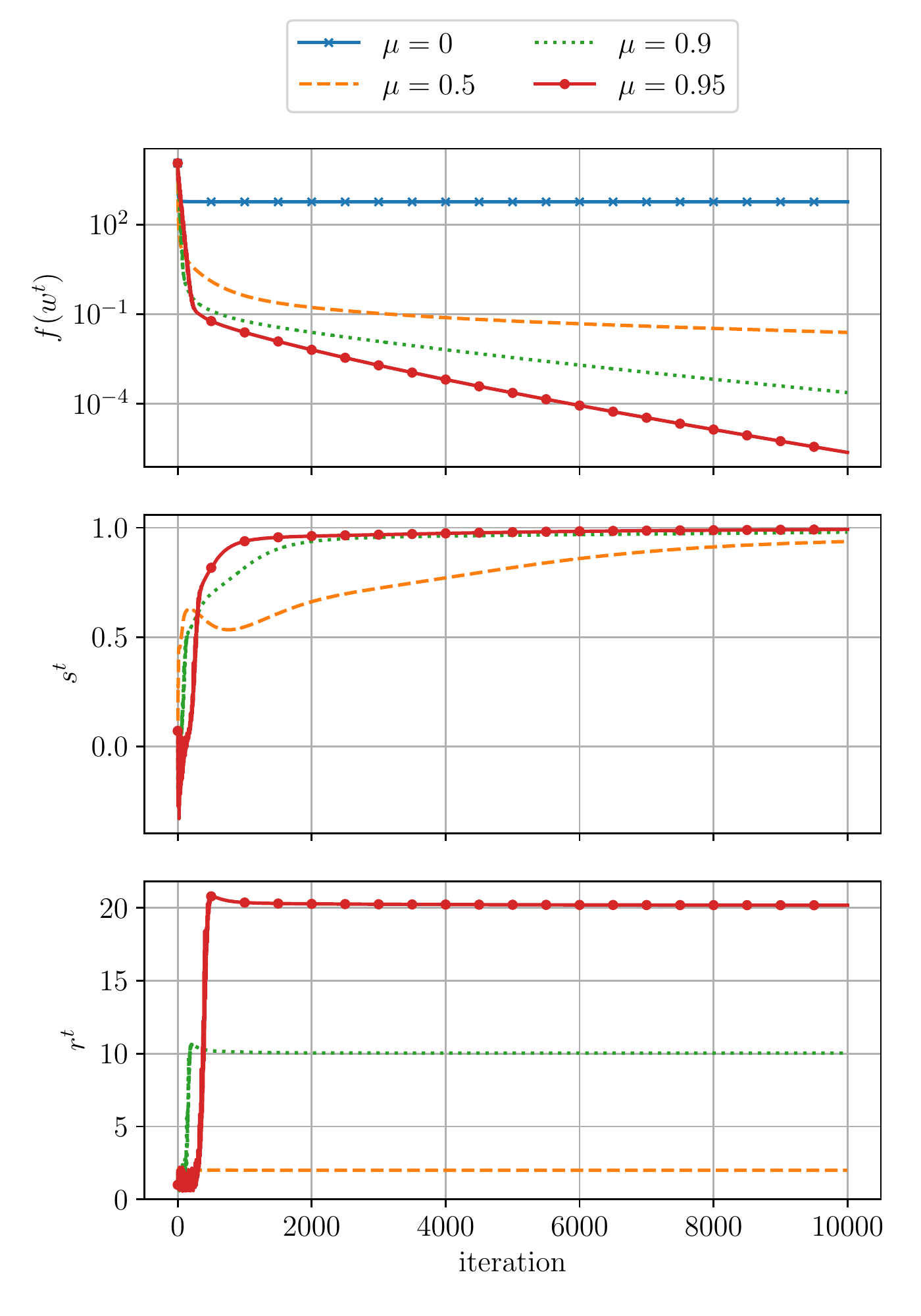}
  \caption{\label{fig:momentum_demo}
Evolution of (top) the value of the function,
 (middle) the alignment $s^t$ and (bottom) the scale $r^t$ while optimizing a quadratic function
 $f(w) = wAw^T$ where $A$ is a random positive semi-definite matrix whose condition
 number is $10^5$.}
\end{figure}

Note that, in order to be helpful in the optimization process, one would expect
the direction of the momentum vector to be correlated with the direction towards
optimum (alignment to be close to $1$), and its scale large enough to be
significant. Indeed, that's the behavior that \textit{provably} emerges in the
context of convex optimization.

\paragraph{Convex function baseline.}

\Cref{fig:momentum_demo} implies that in case of a quadratic convex function,
a higher momentum value results in faster convergence According to the middle
plot, the momentum vector is a strong indicator of the direction towards
the optimum (it quickly goes to $1$).  Also, the scale $r^t$ increases with the
momentum and eventually converges towards $\frac{1}{1-\mu}$, which is what one
would expect when the momentum is indeed accumulating.

\begin{figure}[t!]
  \begin{center}
     \includegraphics[width=1\columnwidth]{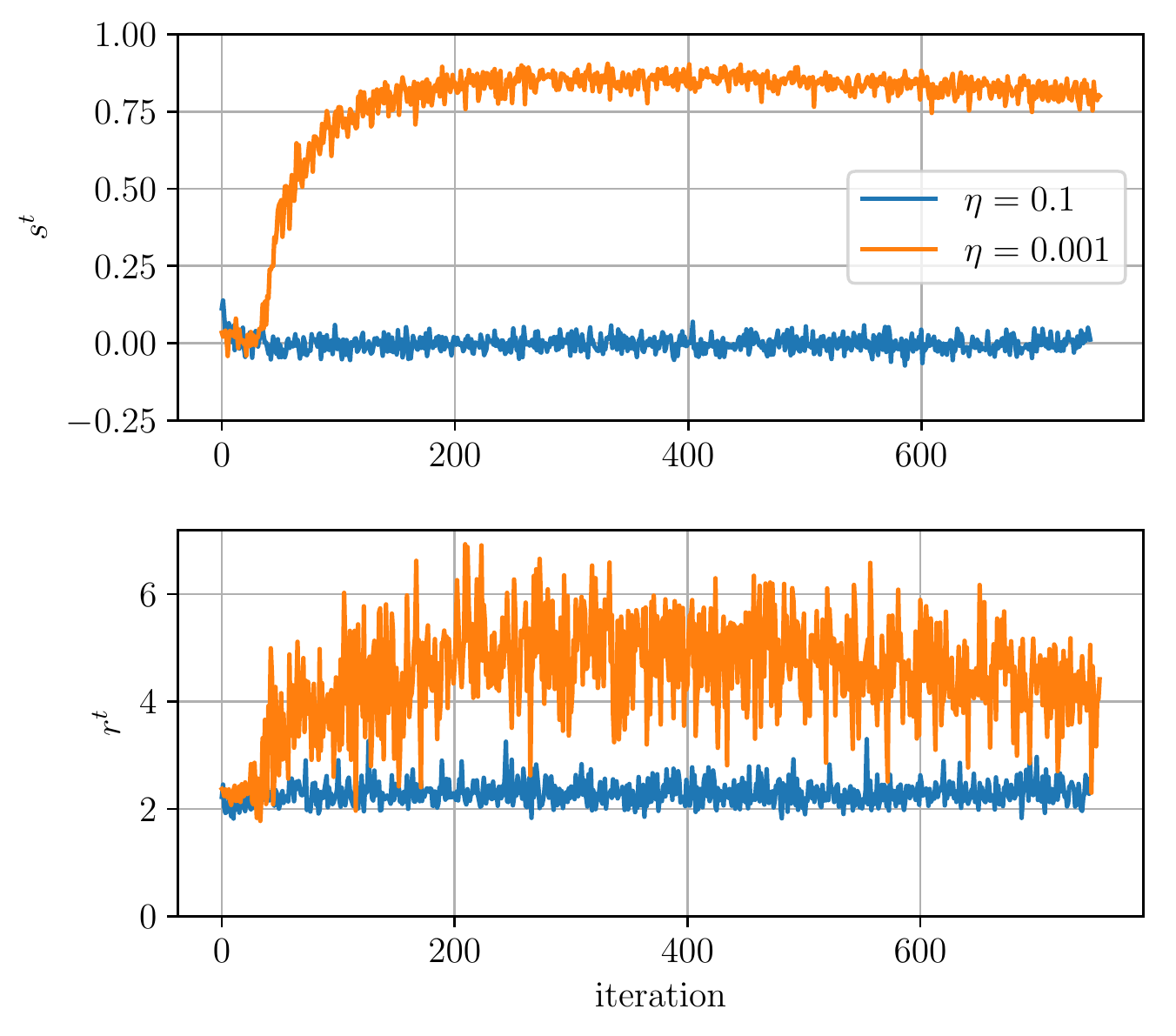}
  \end{center}
 \caption{\label{fig:rs_analysis_sgd_with_low}
 Evolution of the metrics (top) $r^t$ and (bottom) 
 $s^t$  corresponding to the experiment done in \Cref{fig:losses}
 ($\eta = 0.1$)  with $\mu = 0.9$}
\end{figure}

\paragraph{Deep learning setting.}

Now that we saw that the metrics behave as we expect on convex functions, we can
measure them (\Cref{fig:rs_analysis_sgd_with_low}) on the experiment
presented earlier.

In regime \bsr{}, the scale $s^t$ is
oscillating around $0$ and the value of the alignment is very low.  This means
that the momentum vector is nearly orthogonal to the direction of the final
solution and never constitutes a strong signal.
In regime \ssr{}, the momentum vector is able to accumulate more and gives
non-negligible information about the direction towards the point we are
converging to.  

According to \citet{kidambi2018on}, momentum might not be able to cope with the
noise coming from the stochasticity of SGD. While it is plausible,
experiments in appendix~\autoref{sec:full_gradients} using full gradients instead of
mini-batches show that this noisiness has only minimal impact and that the step
size is the most important factor determining the success of momentum.

This leads to the following informal argument: with small step-sizes, the
trajectory is unable to escape the current basin of attraction. The region is
``locally convex'' and, as a consequence, allows the momentum vector to point
towards the same critical point during the optimization process, thus helping to
speed up optimization. On the other hand, high learning rates penalize the
effect of momentum. The steps taken at each iteration are large
enough to escape the current basin of attraction and enter a different basin
(therefore optimizing towards a different local minimum). As this happens, the
direction approximated by the momentum vector points to different critical
points during the course of optimization. Thus, at some iteration, momentum
steers the trajectory towards a point that is not reachable anymore.  This
hypothesis also explains why in the top plot of \Cref{fig:losses}, we see
momentum struggling more than vanilla SGD: often, the momentum vector, orthogonal, is
completely disagreeing with the gradient and slows down convergence.

\subsection{Generalization perspective}
\label{sec:generalization}

If our objective is to minimize the loss, training in the small-step
regime \ssr{} is simpler and faster. Indeed, as we saw in \Cref{fig:losses},
it was two times faster to reach a loss of $3 \times 10^{-4}$. It is therefore
natural to ask: \textit{Why do we even spend some time in the high learning
    rate regime?} In deep learning, the loss is only a surrogate of our real
objective: testing accuracy. It turns out that training only in the second
regime, while it is fast, leads to very sharp minimizers. This is a phenomenon
similar to what was described in~\citet{keskar2017on} in the context of the
batch size.

The relationship between learning rate and generalization has already been
studied in the
past~\citep{li2019,hoffer2017train,keskar2017on,jiang2020,keskar2017on,jiang2020}.  However, it seems that what truly defines the regime we
are in is not the learning rate itself, but the actual step size.\footnote{This is the
reason why we prefer the term \textit{large} and \textit{small steps} regimes
as it is possible to make large steps with a small learning rate if momentum is
large enough.}

Momentum, as defined in \Cref{sec:background}, increases the size of the step
we actually take at each iteration of SGD. As we saw in
\Cref{sec:optimization}, it does not seem to be able to speed up the
optimization process. However, it is easy to find parameters where increasing
momentum improve generalization. In this paper, we demonstrate that in the
large-step size regime \bsr{}, momentum solely boosts the step size.

\begin{figure}[h!]
\includegraphics[width=\columnwidth]{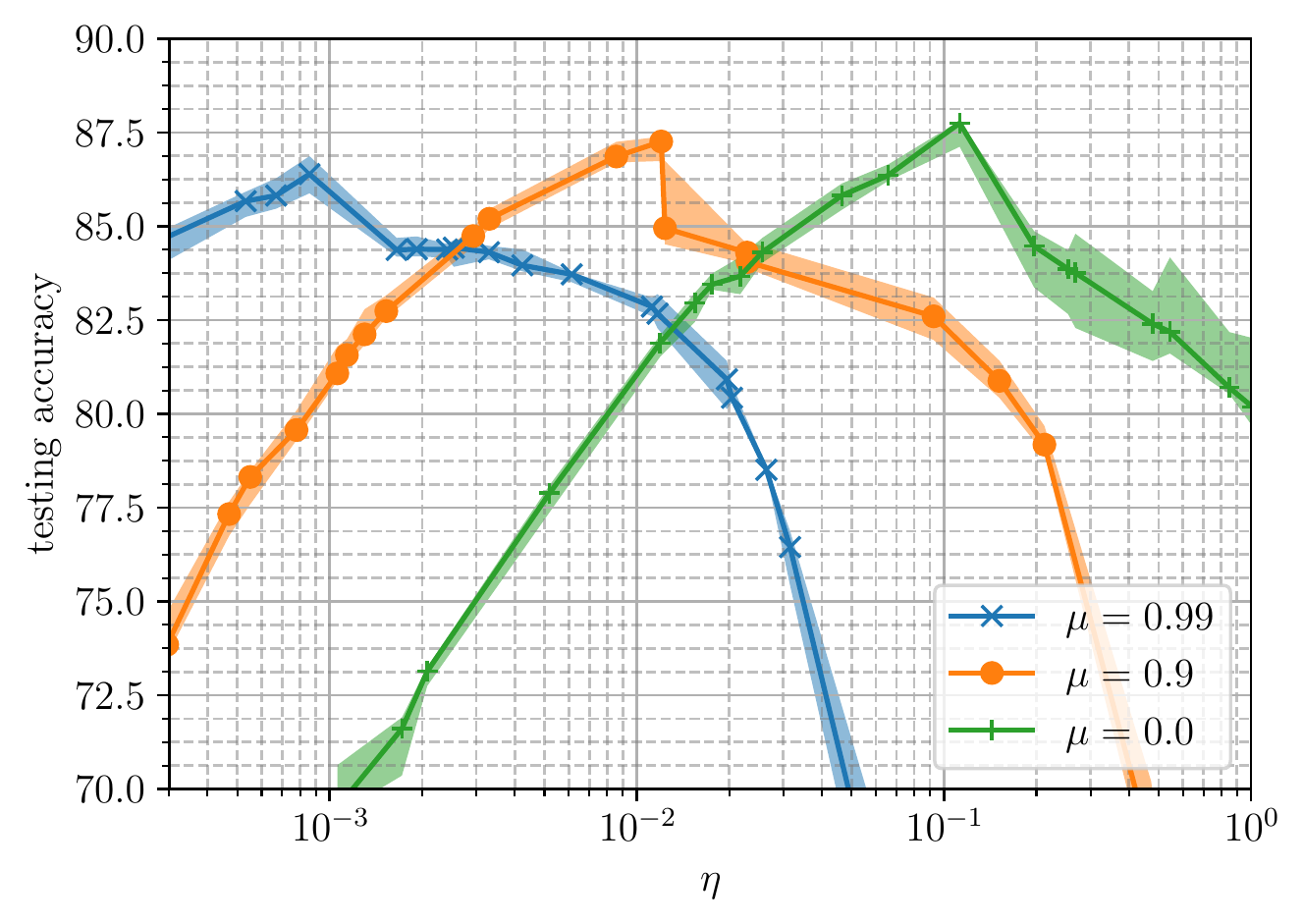}
 \caption{\label{fig:exp9-best-tacc-vs-lr}Testing accuracies obtained for various
 learning rates and three different values of $\mu$.
 \texttt{VGG-13-BN} was trained using SGD.}
\end{figure}

Indeed, assuming that $\frac{||g^{t+1}||_2}{||\nabla f(w^t)||_2}$ does not fluctuate
much during training and can be approximated by a constant, then we
can simulate the increase in step size implied by momentum just by
using a higher learning rate (Figure \ref{fig:exp9-best-tacc-vs-lr}).

\begin{figure}[h!]
\includegraphics[width=\columnwidth]{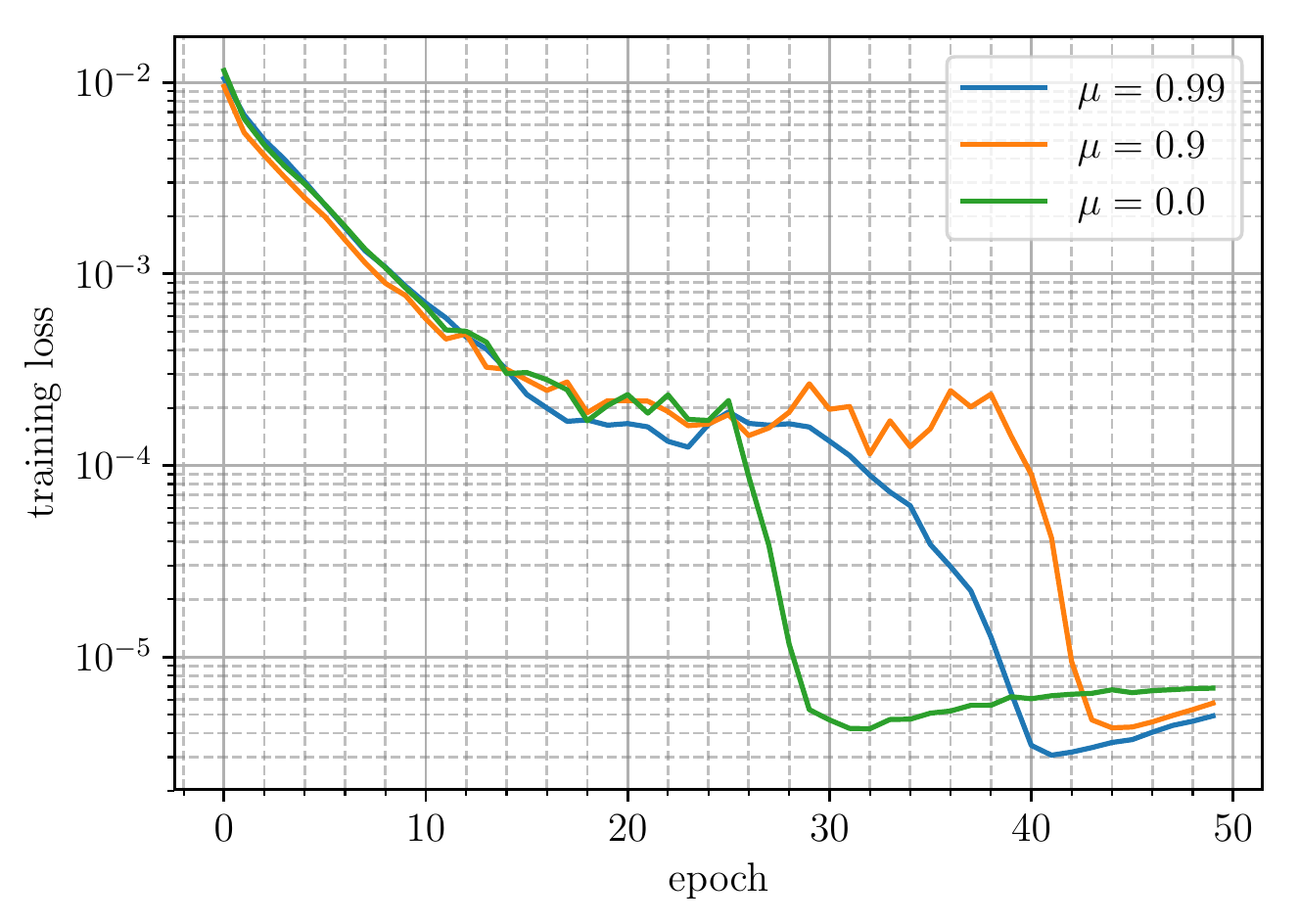}
 \caption{\label{fig:exp9-best-curves}Loss curves of the best learning rate for each
 momentum value in \Cref{fig:exp9-best-tacc-vs-lr}.}
\end{figure}

\Cref{fig:exp9-best-tacc-vs-lr} indicates that the generalization ability is
dictated by the size of the steps taken rather than the learning rate itself.
For three different momentum intensities, we can observe the same pattern
repeating. Reductions in momentum appear to be compensated by increasing the
learning rate.  The three curves, albeit shifted, are surprisingly similar.
They even exhibit the same drop just after their respective optimal learning
rate. Additional experiments made on other architectures and datasets were
performed to rule out the hypothesis that these results are problem specific.
Results are presented on  \Cref{fig:full_exp_9}. Also in appendix
\autoref{sec:momentum_lr_equiv}, we explore in more detail the equivalence of
pairs of learning rate and momentum values.

Finally, \Cref{fig:exp9-best-curves} shows the evolution of the loss during
training for the bets performing learning rate of each momentum value considered
in \Cref{fig:exp9-best-tacc-vs-lr}. It is clear that momentum had no impact
here as the trajectories are oddly similar. There is no evidence that the
convergence was improved at all. The only difference that we can observe is that
each model reached $10^{-4}$ at a different time, but it does not seem to be
linked to the intensity of momentum. Moreover, they all reach very similar
losses at the end of training.

\section{Towards new learning rate schedules}
\label{sec:schedules}

As we characterized these two very distinct training regimes, it is tempting to
experiment with a ``stripped down'' schedule that consists of two completely
different phases; For each one we use an algorithm \textit{individually} tuned
to excel in a particular task. The first one has to be SGD as it provides good
generalization to the model. The second can be \textit{any} algorithm able to
minimize the loss quickly. To stay consistent with the previous experiments we
pick here SGD with momentum but we believe that many algorithms would perform
similarly or better.  Especially, fast algorithms that have been criticized for
their poor generalization ability like K-FAC~\citep{MartensG2015} and
L-BFGS~\citep{LiuN1989} could be perfect candidates.  First, we will appraise
the benefits of having radically different momentum values for the two phases.
Secondly, we will evaluate the performance of this two step approach in
comparison to the more elaborate three phase training schedule.

\subsection{Decoupling momentum}
\label{sec:decoupling_momentum}

We believe that, even if researchers do search for the best momentum value,
unlike learning rate, they assume that it stays constant. For example,
in~\citet{goyal17} and~\citet{shallue2018measuring}, a large amount of schedules
are compared; yet momentum never change over the course of training. However,
as we saw, the two regimes are wildly different. This is why we suggest
isolating the two regimes in the two tasks, and optimize them individually.

It turns out that with the appropriate learning rate, using momentum in the
first phase has no observable impact on the performance of the models. However,
having a larger momentum (again with an appropriate change in learning rate) is
beneficial in the second phase as it inceases the final testing accuracy under
the same budget.

\begin{figure}[h!]
\centering
\includegraphics[width=\columnwidth]{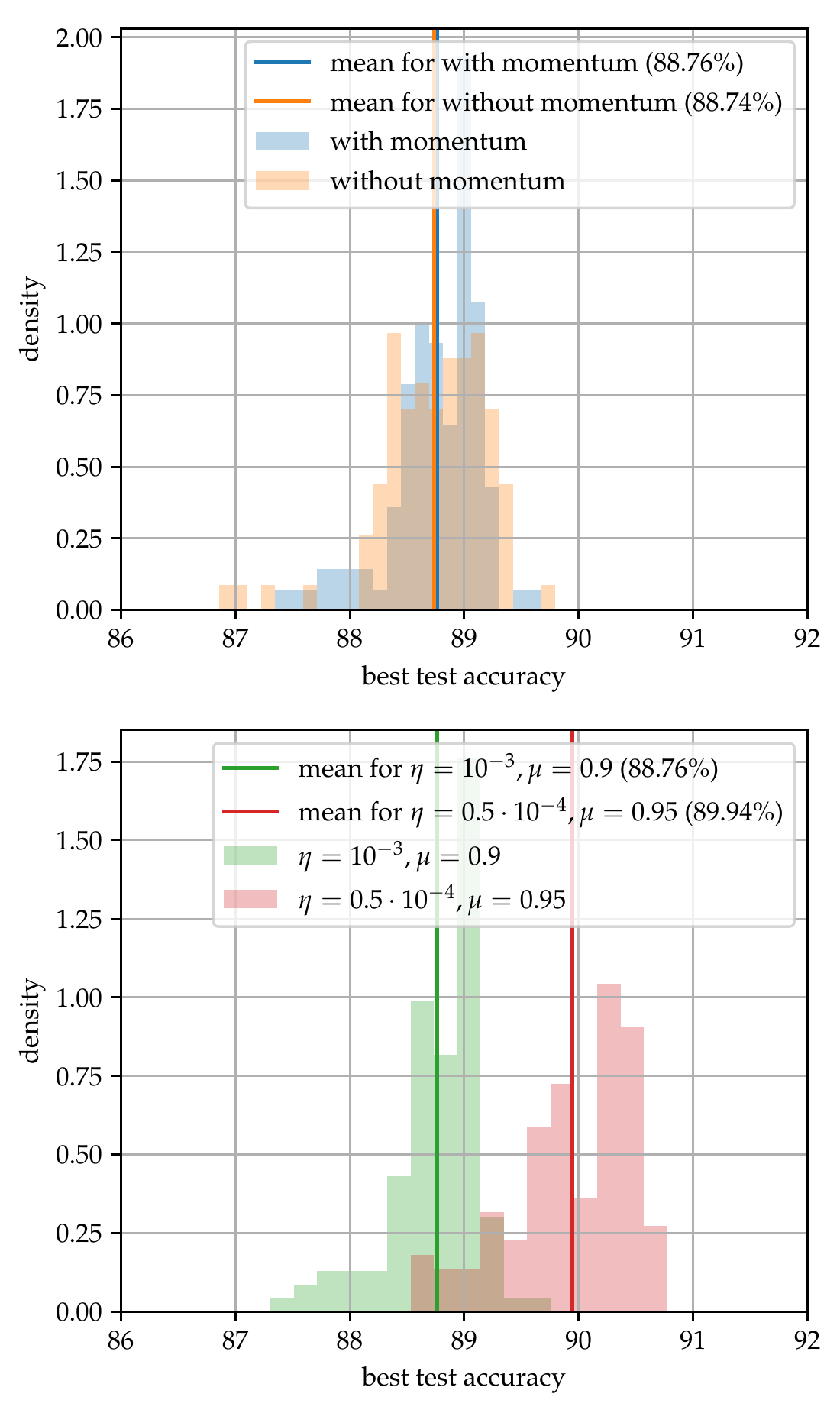}
    \caption{Impact on the distribution of testing accuracies when using
        different values momentum in the training phases, (top) is for the
        first phase and (bottom) is for the second.}
         \label{fig:exp_11_generic_vs_specialized}
\end{figure}

In order to control for the parameters, we trained multiple models and
randomly picked the \textit{transition epoch}, epoch at which we switch from an
algorithm to another. We display the
distribution of testing accuracies obtained on
\Cref{fig:exp_11_generic_vs_specialized}\footnote{Results for different
    second phase algorithms are available in the appendix on
    \autoref{fig:exp_11_first_phase_momentum_useless}.}. On the top plot we
see that two distributions are the same (for a fixed
second phase). On the bottom one, however, a more aggressive momentum
associated with a smaller learning rate, on average, outperforms the ``classic''
parameters.

We previously observed that disabling momentum has to be accompanied by a
corresponding increase in learning rate. To find such a learning rate we used random
search and took the one that had a training loss curve that matched the
baseline as closely as possible for the first 50 epochs(more details about this
procedure in appendix \autoref{sec:momentum_lr_equiv}).

\subsection{Performance of the two phases schedule}

Comparing against the popular, three-step schedule, we find that two truly
independent phases can perform similarly or better. This suggests that complex
schedules are not necessary to train deep neural networks.

We evaluate this schedule on two datasets: \texttt{CIFAR-10} and
\texttt{ImageNet}~\citep{russakovsky2015imagenet}. For the former, we sampled many
transition points and took the median over equally sized bins. For the latter,
because it is particularly expensive, only a few transition points were hand
picked. For \texttt{CIFAR-10}, we used the same parameters as in
\Cref{sec:decoupling_momentum}. For \texttt{ImageNet}, learning rates and
momentum values were hand picked, as optimizing them would have been
prohibitively costly.

\begin{figure}[p]
\centering
\includegraphics[width=\columnwidth]{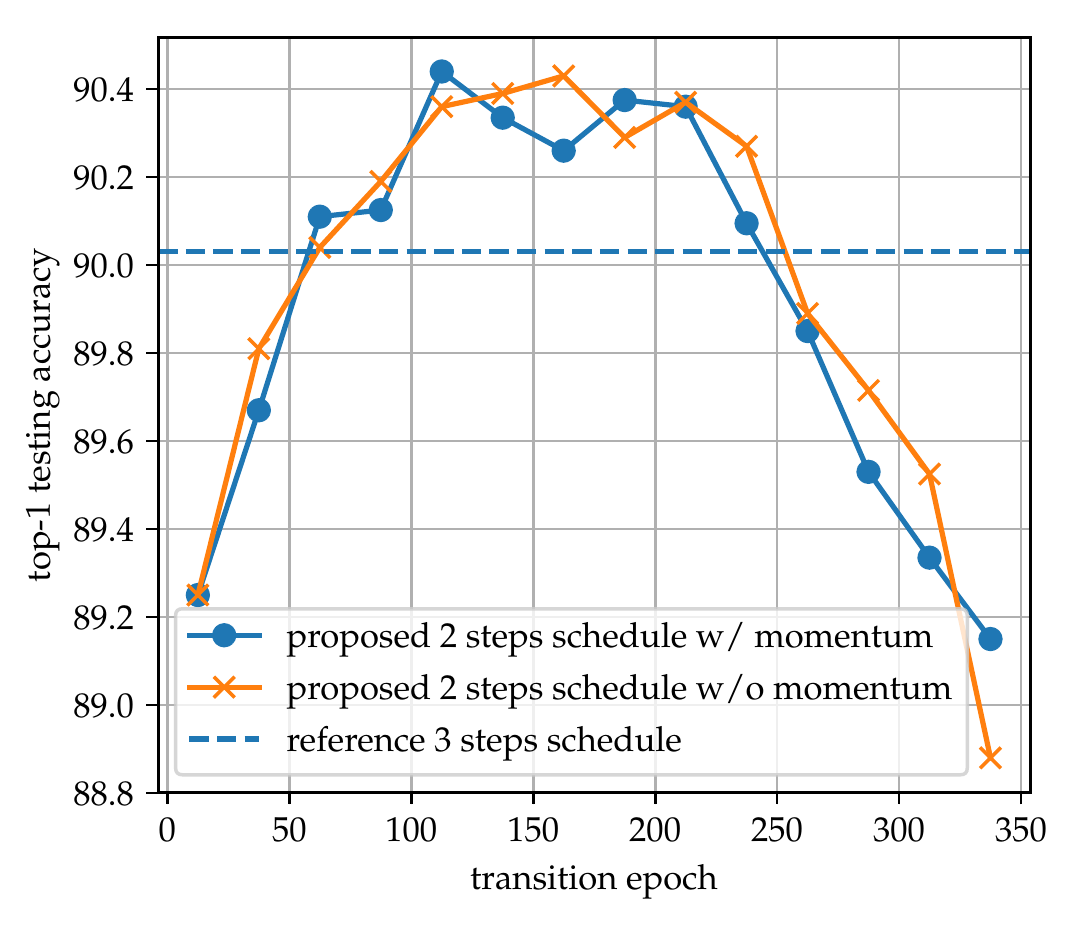}
    \caption{Evolution of the testing accuracy in function of the transition epoch
    for our proposed simplified two steps schedule on \texttt{CIFAR-10}. We
    present our schedule with and without momentum in the first phase to
    emphasise its lack of influence on the results.}
         \label{fig:exp_11_distribution}
\end{figure}

\begin{figure}[p]
\centering
\includegraphics[width=\columnwidth]{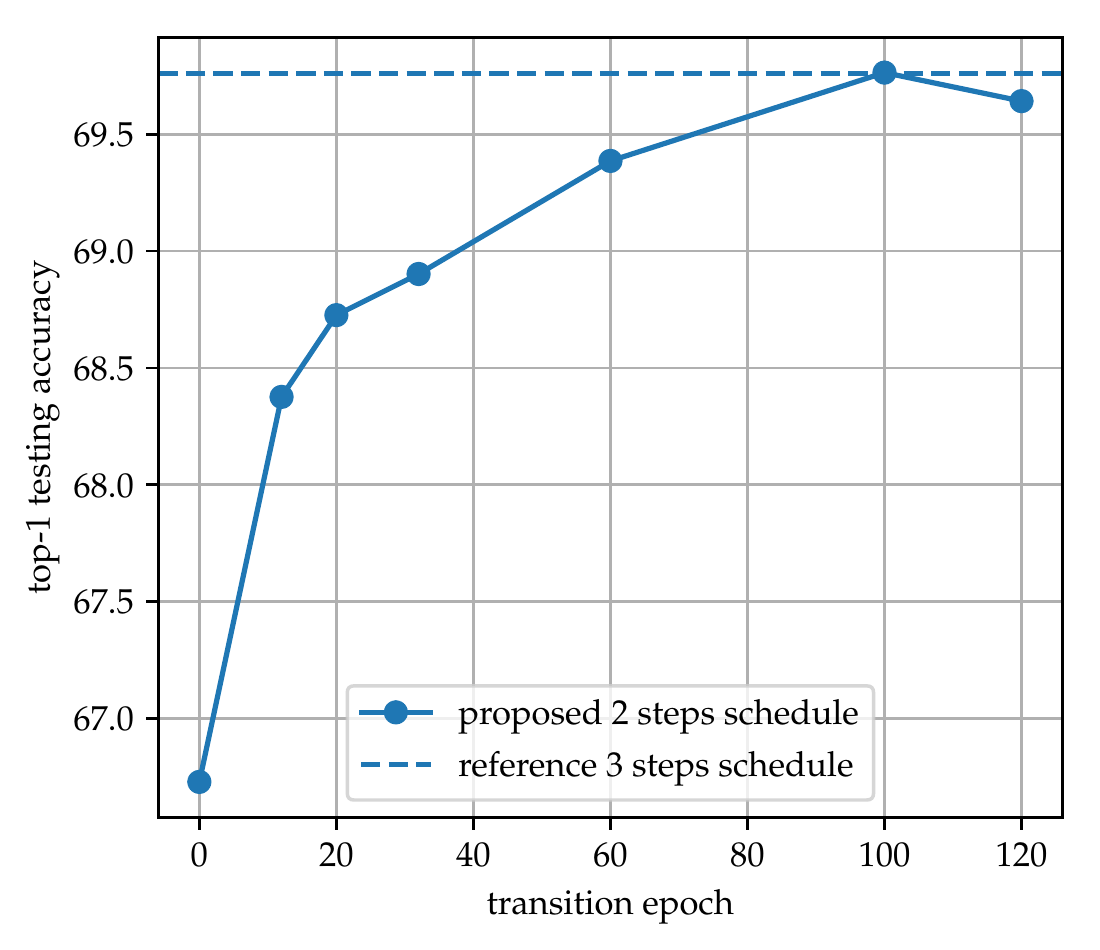}
    \caption{Evolution of the testing accuracy in function of the transition epoch
    for our proposed simplified two steps schedule on \texttt{ImageNet}.}
         \label{fig:exp_imagenet_training}
\end{figure}

Performance as a function of the \textit{transition epoch} is shown on
\Cref{fig:exp_11_distribution} and \Cref{fig:exp_imagenet_training}. In
both cases, our schedule outperforms or matches the three stages schedule for at
least ones value of the \textit{transition epoch}. For \texttt{CIFAR-10}, we also
considered enabling momentum in the first phase\footnote{with the appropriate
change in learning rate}. As our previous experiment would suggest, the two
configurations appear equivalent.

\section{Related work}
\label{sec:related}

The older and more popular multiple step learning rate schedules probably
originates from the practical recommendations found
in~\citet{Bengio2012}. \citet{bottou18optimization} provides a theoretical argument
that support schedules with decreasing learning rate.

More recently, \citet{Smith17} introduced the cyclic learning rate that consist
in a sequence of linear increase and decrease of the learning rate where the
high and low values correspond to what we named the large and small step
regimes.  Soon after, \citet{SmithT17} concludes that a single period of that
pattern is sufficient to obtain good performance and the schedule is named
\texttt{1-cycle}. Similarly to the cycling learning rate schedule,
\citet{loshchilov17sgdr} present SGDR, a schedule with sudden jumps of learning
rate similar to the restarts found in many gradient free optimization
techniques.

The learning rate is not the only parameter that has been considered to change
over time. For example, ~\citet{smith18} and~\citet{goyal17} both had success
varying the size of the batch size. This is aligned with our recommendation that
every hyper-parameter should be optimized for each phase of training.

The impact of large learning rates and generalization has received a lot of
attention in the past. The predominant hypothesis is that is acts as
regularizer~\citep{li2019,hoffer2017train}. It is believed that it either
promotes flatter minima~\citep{keskar2017on,jiang2020} or increase the amount of
noise during training~\citep{mandt17,smith18}.

\section{Conclusion}
\label{sec:conclusion}

In this paper, we studied the properties of the two regimes of deep network
training. The large-step one is typically found in the early stages of
training\footnote{Cyclic learning rates and the schedule used in \citet{goyal17}
are example of exceptions.} and the small-step tends to ends training.

Our investigations show that optimization in the large
step-size regime does not follow training patterns typically expected
in the convex setting: the evolution of the loss is very noisy and we reach a
solution far from the optimal one. In this regime, the benefits of momentum are nuanced:
It seems that any gain that it offers can be compensated by a
corresponding increase in learning rate.

The small step-size regime seems fundamentally different: we obtain a lower
loss, faster and smoothly, but solutions generalize poorly. In this case,
momentum can greatly speed up the convergence---as it does in the convex case.

The intensity of momentum and, more generally, the optimization
algorithm used are typically considered during hyper-parameter search.
However, they are always kept constant over the whole training. This restrict the
search space drastically because we are unable to tailor them to the
different training regimes we encounter. By separating the two regimes into two
distinct problem we might be able to obtain better model and/or train them
faster.

Indeed, we demonstrate that a simple schedule consisting of only two stages, the
first one being SGD with \textit{no} momentum and the second of SGD with
a value of momentum larger than usual can be competitive with state of the art
learning rate schedules. This opens up the possibility for development of new
training algorithms that are specialized in only one regime.
This might also let us leverage second order methods---usually criticized for
the poor generalization performance--- in the second phase of training.

\FloatBarrier
\bibliographystyle{unsrtnat}
\bibliography{custom,mendeley,madrylab,main}

\clearpage
\begin{appendices}
\section{Full gradient experiments}
\label{sec:full_gradients}

In this appendix we reproduce some of the experiments made in
\Cref{sec:optimization} and \Cref{sec:generalization} using full gradients
instead of SGD to rule out the possibility that the stochasticity is the cause
of the inability of momentum to build up.

\begin{figure}[p]
 \centering
 \includegraphics[width=1\columnwidth]{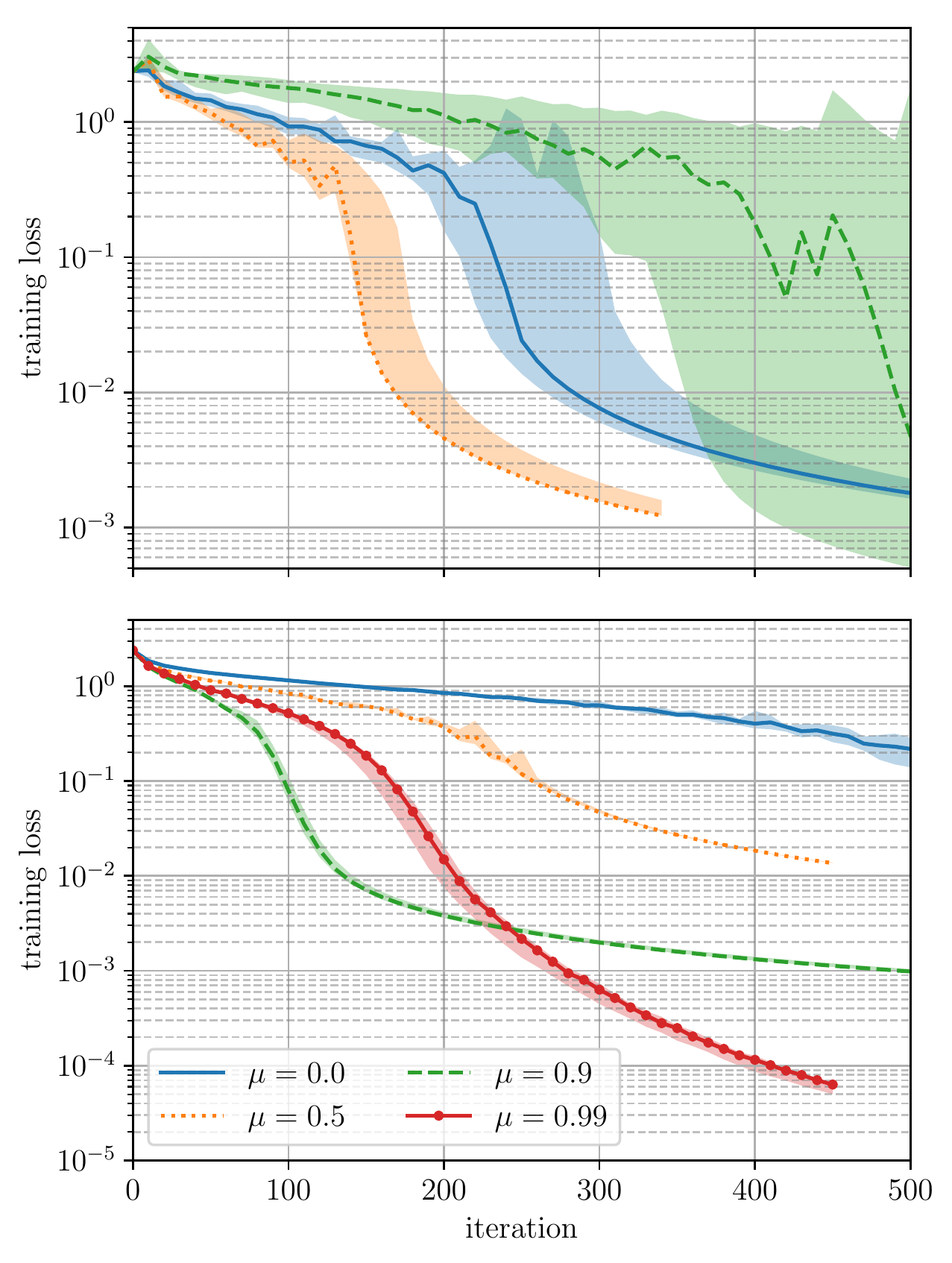}
 \caption{\label{fig:exp2-loss-high-low}
 Evolution of the training loss using full gradients with different momentum values on \texttt{CIFAR-10} and \texttt{VGG-13-BN} with $\eta=0.1$ for (a) and $\eta = 0.01$ for (b).}
\end{figure}

\begin{figure}[p]
\centering
\includegraphics[width=1\columnwidth]{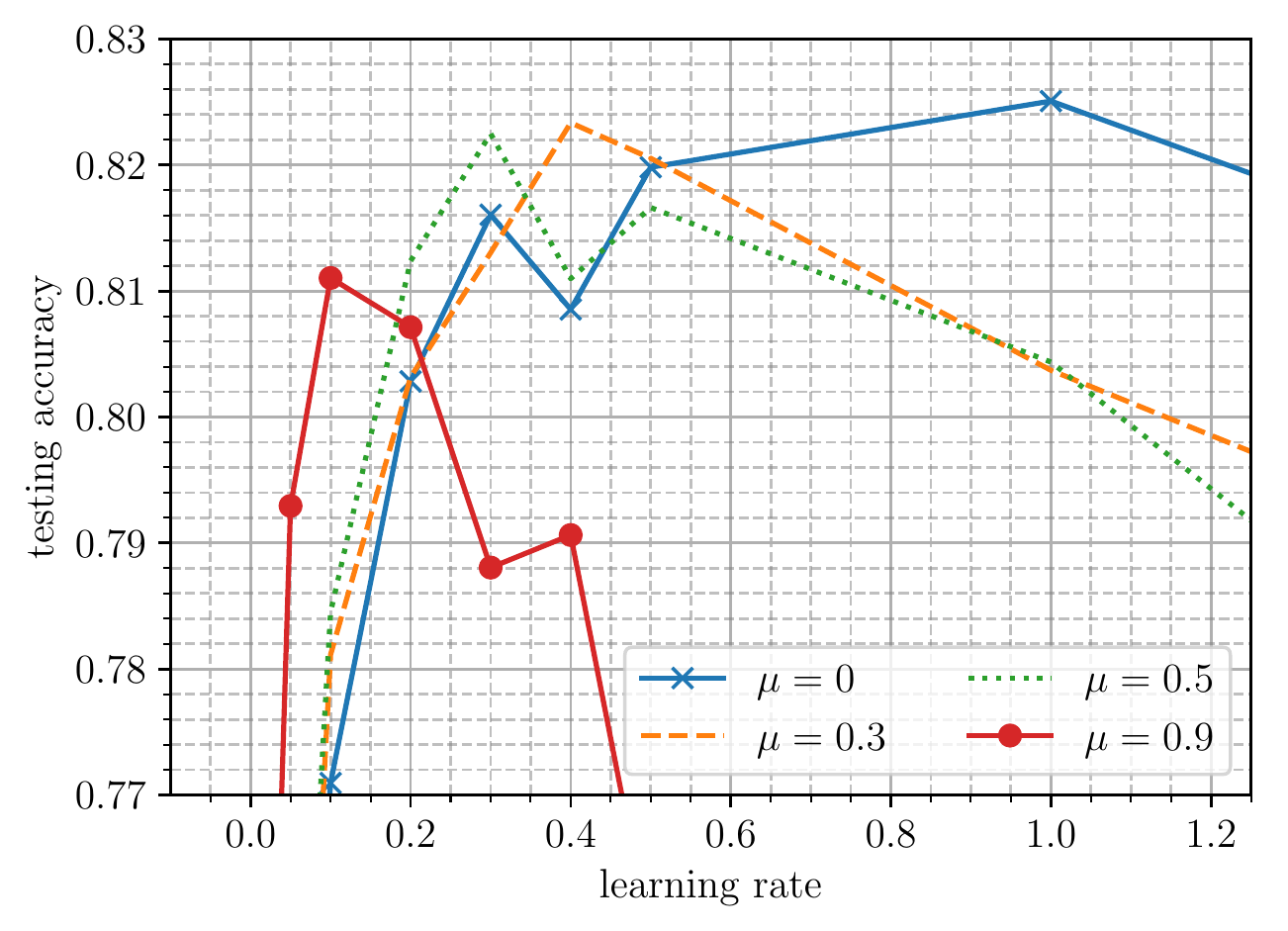}
 \caption{\label{fig:exp2-tacc_vs_lr_gd}
 Testing accuracy after 50 epochs in function of the learning rate for different
 values of $\mu$. \texttt{VGG-13-BN} model was
 used and trained using GD and no weight decay.}
\end{figure}

\Cref{fig:exp2-loss-high-low} and \Cref{fig:exp2-tacc_vs_lr_gd} presents similar results to
\Cref{fig:losses} and \Cref{fig:exp9-best-tacc-vs-lr} respectively.

\section{Momentum learning-rate equivalence in the large step regime}
\label{sec:momentum_lr_equiv}

\begin{figure}[]
 \centering
 \includegraphics[width=1\columnwidth]{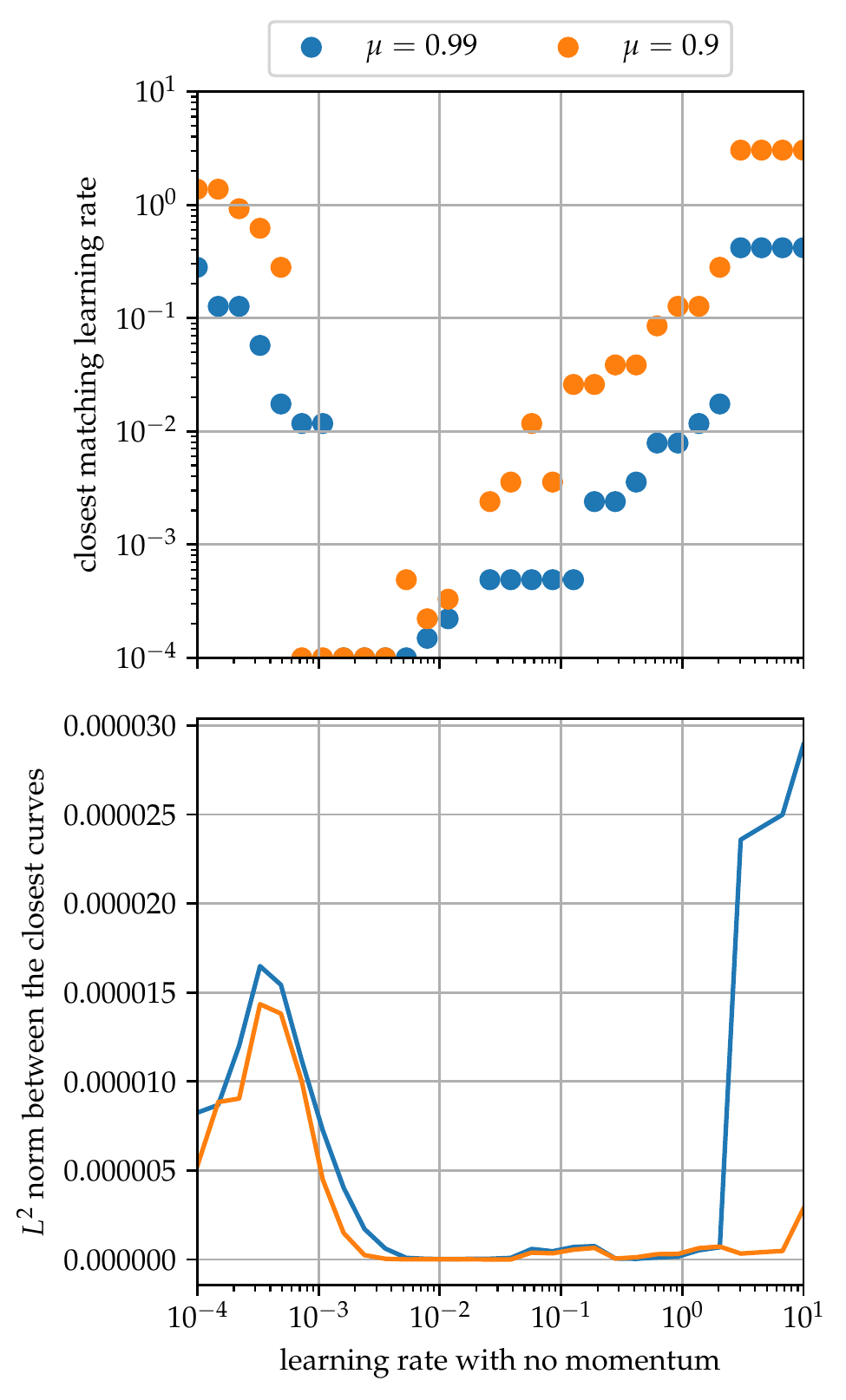}
 \caption{\label{fig:grid_search_summary}Best equivalent learning rate (top) and
 corresponding $L^2$ distance between the loss curves for different values of
 momentum.}
\end{figure}

To evaluate in more detail the relationship that ties the learning and the
momentum together, we designed the following experiment:
\begin{enumerate}
\item Train models for \texttt{CIFAR-10} for 50 epochs with a wide range of learning rates and
    three different momentum intensities: $0$, $0.9$ and $0.99$.
\item For each configuration with $\mu=0$ we find the corresponding two
    configurations with $\mu=0.9$ and $\mu=0.99$ that match the training loss
    the best in $L^2$ norm.
\item We report the corresponding matching learning rate and norm between the
    curves on \Cref{fig:grid_search_summary}.
\end{enumerate}

We see that for any learning rate between $5\cdot10^{-3}$ and $2$ it is
possible to find an equivalent learning rate with an almost identical behavior.
Moreover, the relation between equivalent learning rates seems to be linear.

\begin{figure*}[]
 \centering
 \includegraphics[width=1.6\columnwidth]{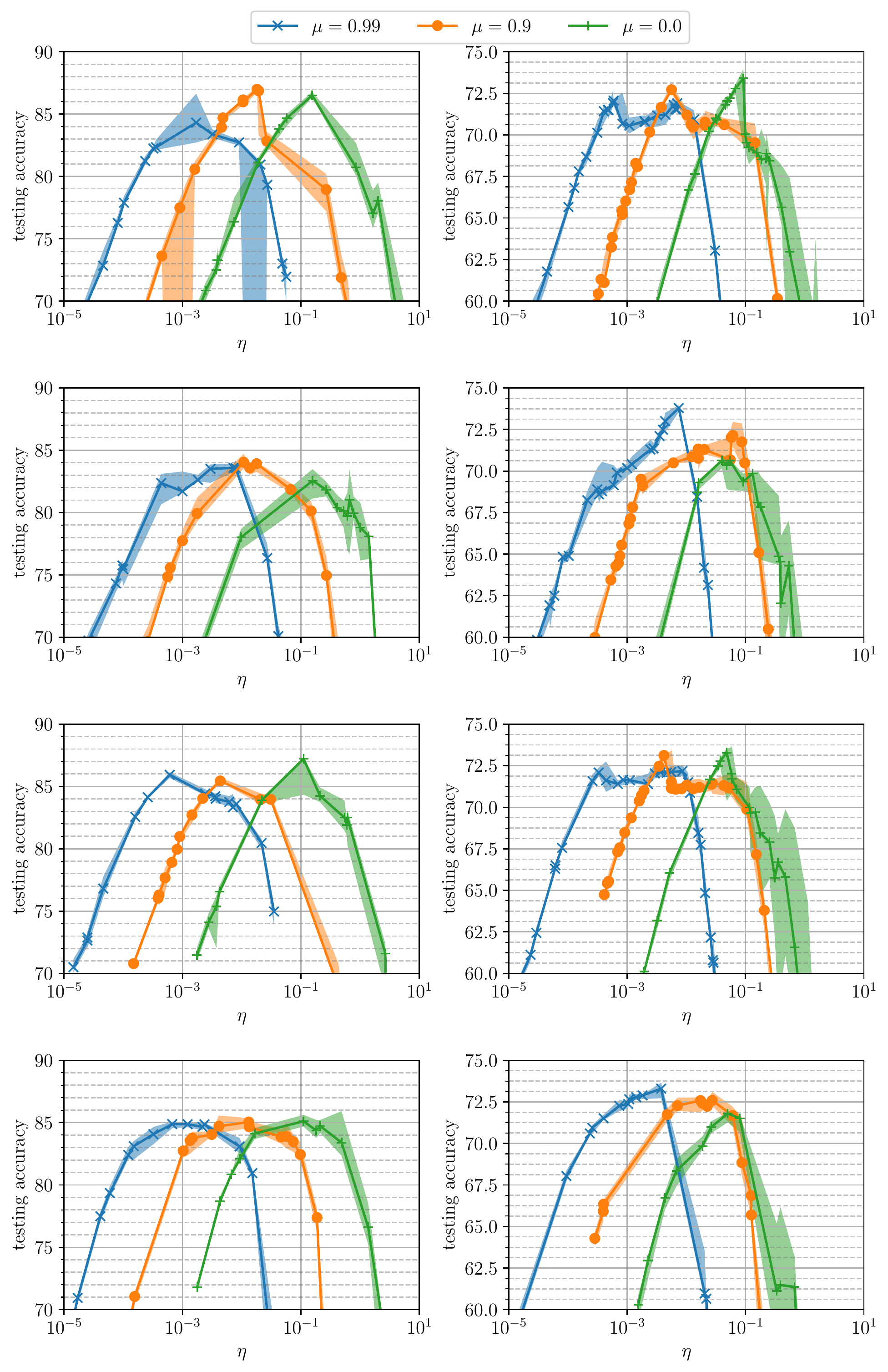}
 \caption{\label{fig:full_exp_9}
Testing accuracies obtained for various learning rates and three different
values of $\mu$.  \texttt{VGG-13-BN} was trained using SGD. Models used are the
same for each row and are, from top to bottom: \texttt{ResNet18},
\texttt{ResNet50}, \texttt{VGG13}, \texttt{VGG19}. (left) column shows
\texttt{CIFAR-10} and (right) \texttt{CINIC-10}.
}
\end{figure*}

\begin{figure*}[]
 \centering
 \includegraphics[width=1.4\columnwidth]{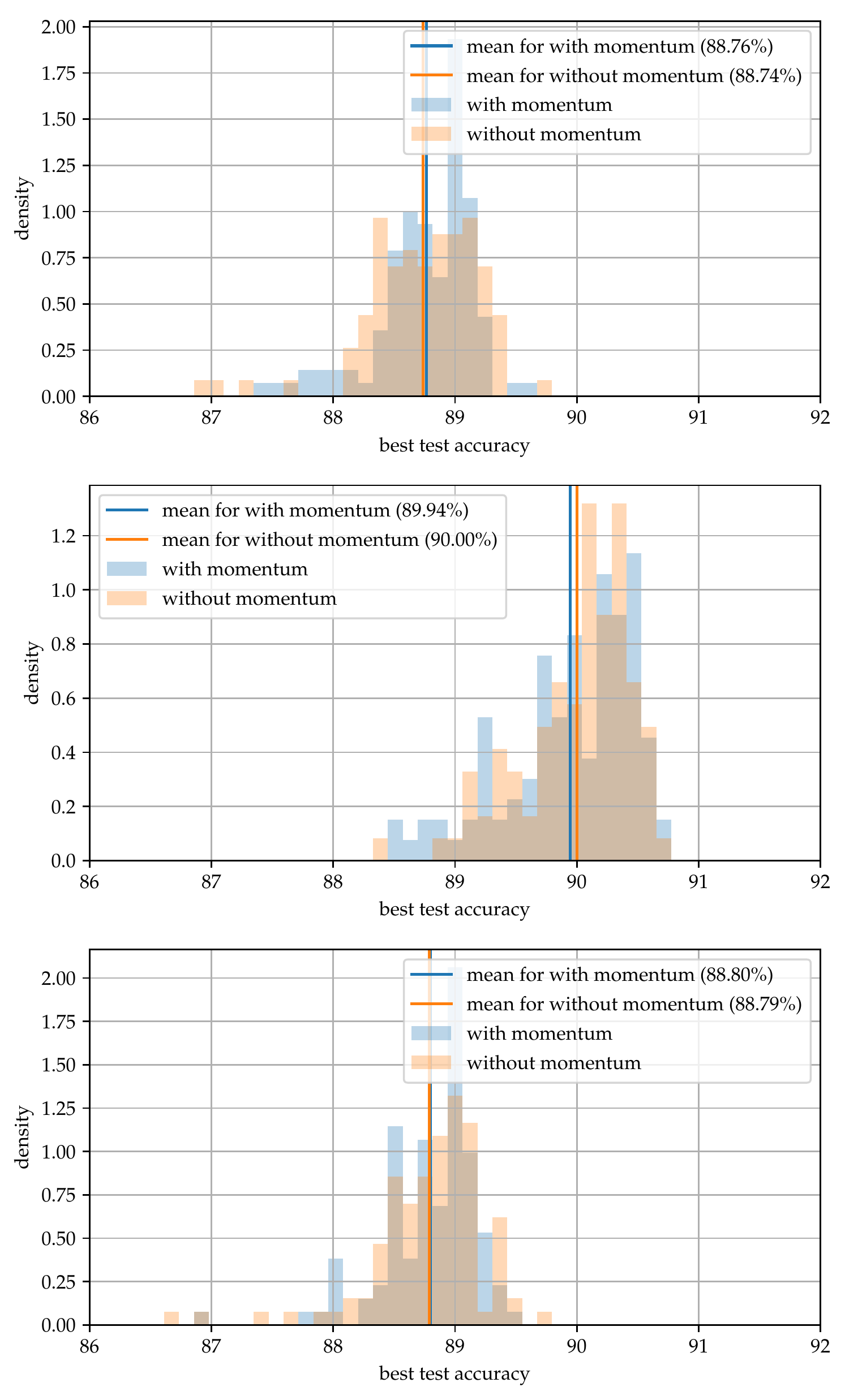}
 \caption{\label{fig:exp_11_first_phase_momentum_useless}Distribution of test
     accuracies with and without momentum in the first phase for different
     second phase algorithms: (top) classic, (middle) reduced learning rate,
     (AdamW)}
\end{figure*}

\section{Experiment details}
\subsection{Shared between all experiments}
The details provided in this section are valid for every experiment unless
specified otherwise:

\begin{itemize}
\item \textbf{Programming language:} Python 3
\item \textbf{Framework} PyTorch 1.0
\item \textbf{Dataset} \texttt{CIFAR-10}~\citep{krizhevsky2009learning}
\item \textbf{Batch size:} $256$
\item \textbf{Weight decay:} $10^{-4}$
\item \textbf{Per channel normalization} Yes
\item \textbf{Data augmentation:} \begin{enumerate}
\item Random Crop
\item Random horizontal flip
\end{enumerate}
\end{itemize}

\subsection{Experiment visible on \Cref{fig:losses} and \Cref{fig:rs_analysis_sgd_with_low}}
\begin{itemize}
\item \textbf{Architecture}: \texttt{VGG-13}~\citep{simonyan2015very} with extra
batch norm layers~\citep{ioffe2015batch}.
\item \textbf{Learning rates:} $\eta = 0.1$ and $\eta = 0.001$ for the large and
small steps regime respectively.
\item \textbf{Momentum type:} Heavy ball (non Nesterov)
\item \textbf{Momentum intensities:} $0$, $0.5$ and $0.9$
\end{itemize}

\subsection{Experiment visible on \Cref{fig:momentum_demo}}
\begin{figure*}[p]
\centering
\includegraphics[width=1.5\columnwidth]{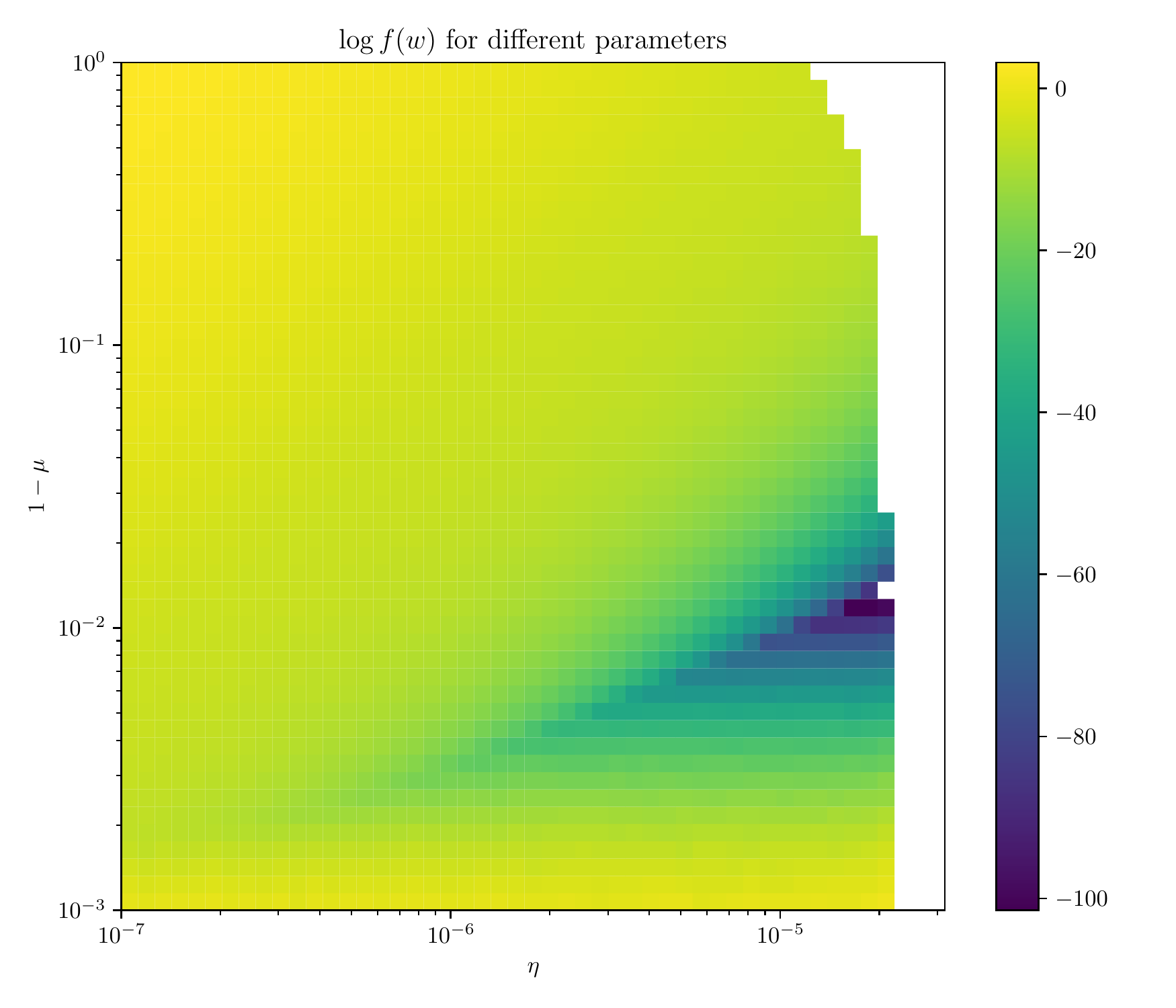}
\caption{\label{fig:momentum_demo_grid_search} Result of the grid search to find
the best learning rate for different momentum intensities. Is displayed $\log f(w)$ after 10000
iterations.}
\end{figure*}

\begin{itemize}
\item \textbf{Function optimized}: $f(w) = wAw^T$
\item \textbf{Iterations}: $10000$
\item \textbf{Properties of $A$}: Fixed positive semi definite random matrix
with eigen values ranging from $1$ to $10^5$.
\item \textbf{Momentum type:} Heavy ball (non Nesterov)
\item \textbf{Momentum intensities:} $0$, $0.5$, $0.9$ and $0.95$
\item \textbf{Learning rates:} They were picked to yield the best performance for
each momentum value. They were obtained using a grid search procedure. Results
of the grid search visible on \Cref{fig:momentum_demo_grid_search}
\item \textbf{Grid search range:} $\eta \in [10^{-7}, 5\times10^5]$ 50 values
equally spaced in $\log$-scale, $1 - \mu \in [10^{-3}, 1]$ 50 values equally
spaced in $\log$-scale.
\end{itemize}

\subsection{Experiment visible on \Cref{fig:exp9-best-tacc-vs-lr} and
\Cref{fig:exp9-best-curves}}

\begin{itemize}
\item \textbf{Architecture}: \texttt{VGG-13}~\citep{simonyan2015very} with extra
batch norm layers~\citep{ioffe2015batch}.
\item \textbf{Momentum type:} Heavy ball (non Nesterov)
\item \textbf{Momentum intensities:} $0$, $0.5$ and $0.9$
\item \textbf{Learning rates:} We performed a random search to find the best for
each each momentum value. We took 20 samples uniformly in log scaled in the
following ranges:
\begin{itemize}
\item $\mu = 0: \eta \in [10^{-3}, 10]$
\item $\mu = 0.9: \eta \in [10^{-4}, 1]$
\item $\mu = 0.99:\eta \in  [10^{-5}, 10^{-1}]$
\end{itemize}
\end{itemize}

\subsection{Experiment vibile at the top of \Cref{fig:exp_11_generic_vs_specialized}}
\begin{itemize}
\item \textbf{Framework} \texttt{PyTorch 0.4.1}
\item \textbf{Architecture}: \texttt{VGG-13}~\citep{simonyan2015very} with extra
batch norm layers~\citep{ioffe2015batch}.
\item \textbf{Batch size:} $256$
\item \textbf{Optimizers} \begin{enumerate} 
\item Phase 1 (with momentum): SGD \begin{enumerate}
    \item Learning rate: $0.1$
    \item Momentum: $0.9$
    \item Weight decay: $10^{-4}$
\end{enumerate}
\item Phase 1 (without momentum): SGD \begin{enumerate}
    \item Learning rate: $0.9236708571873865$
    \item Momentum: $0$
    \item Weight decay: $10^{-4}$
\end{enumerate}
\item Phase 2 (same for the two distributions): SGD with momentum \begin{enumerate}
    \item Learning rate: $0.001$
    \item Momentum: $0.9$
    \item Weight decay: $10^{-4}$
\end{enumerate}
\end{enumerate}

\end{itemize}

\subsection{Experiment vibile at the bottom of \Cref{fig:exp_11_generic_vs_specialized}}
\begin{itemize}
\item \textbf{Framework} \texttt{PyTorch 0.4.1}
\item \textbf{Architecture}: \texttt{VGG-13}~\citep{simonyan2015very} with extra
batch norm layers~\citep{ioffe2015batch}.
\item \textbf{Batch size:} $256$
\item \textbf{Optimizers} \begin{enumerate} 
\item Phase 1 (same for the two distributions): SGD \begin{enumerate}
    \item Learning rate: $0.1$
    \item Momentum: $0.9$
    \item Weight decay: $10^{-4}$
\end{enumerate}
\item Phase 2  SGD \begin{enumerate}
    \item Learning rate: displayed on the legend
    \item Momentum: displayed on the legend
    \item Weight decay: $10^{-4}$
\end{enumerate}
\end{enumerate}

\end{itemize}

\subsection{Experiment visible on \Cref{fig:exp_11_distribution}}
\begin{itemize}
\item \textbf{Framework} \texttt{PyTorch 0.4.1}
\item \textbf{Architecture}: \texttt{VGG-13}~\citep{simonyan2015very} with extra
batch norm layers~\citep{ioffe2015batch}.
\item \textbf{Batch size:} $256$
\item \textbf{Data augmentation:} None
\item \textbf{Optimizers} \begin{enumerate} 
\item Phase 1: SGD \begin{enumerate}
    \item Learning rate: $0.9236708571873865$
    \item Momentum: $0$
    \item Weight decay: $10^{-4}$
\end{enumerate}
\end{enumerate}
\begin{enumerate}
\item Phase 2: SGD with momentum \begin{enumerate}
    \item Learning rate: $0.005$
    \item Momentum: $0.95$
    \item Weight decay: $10^{-4}$
\end{enumerate}
\end{enumerate}
\item \textbf{Momentum type:} Classic
\item \textbf{Reference testing accuracy:} Median over multiple training runs
    that we ran ourself wit the same parameters except for the learning rate
    schedule. The default three stages schedule was used with a constant
    $\mu=0.9$.
\end{itemize}

\subsection{Experiment visible on \Cref{fig:exp_imagenet_training}}
\begin{itemize}
\item \textbf{Framework} \texttt{PyTorch 0.4.1} + \texttt{Robustness 1.1}
\item \textbf{Architecture}: \texttt{ResNet-18}~\citep{he2016deep} 
\item \textbf{Batch size:} $256$
\item \textbf{Data augmentation:} \begin{enumerate}
\item Random crop to size 224
\item Random horizontal flip
\item Color Jitter
\item Lighting noise
\end{enumerate}
\item \textbf{Optimizers} \begin{enumerate} 
\item Phase 1: SGD \begin{enumerate}
    \item Learning rate: $1$
    \item Momentum: $0$
    \item Weight decay: $10^{-4}$
\end{enumerate}
\end{enumerate}
\begin{enumerate}
\item Phase 2: SGD with momentum \begin{enumerate}
    \item Learning rate: $10^{-4}$
    \item Momentum: $0.995$
    \item Weight decay: $10^{-4}$
\end{enumerate}
\end{enumerate}
\item \textbf{Momentum type:} Classic
\item \textbf{Reference testing accuracy:} We used the value availble here:
    \url{https://pytorch.org/docs/stable/torchvision/models.html} the day of
    submission.
    \end{itemize}

\end{appendices}

\end{document}